\documentclass[final]{cvpr}

\usepackage{times}
\usepackage{epsfig}
\usepackage{graphicx}
\usepackage{adjustbox}
\usepackage{amsmath}
\usepackage{amssymb}

\usepackage[utf8]{inputenc} % allow utf-8 input
\usepackage[T1]{fontenc}    % use 8-bit T1 fonts
\usepackage{url}            % simple URL typesetting
\usepackage{booktabs}       % professional-quality tables
\usepackage{amsfonts}       % blackboard math symbols
\usepackage{nicefrac}       % compact symbols for 1/2, etc.
\usepackage{microtype}      % microtypography
\usepackage{bbm}            % For indicator function.
\usepackage{amsmath}
\usepackage{amssymb}
\usepackage{bm}
\usepackage{graphicx}
\graphicspath{ {./images/} }
\usepackage{multirow}
\usepackage{colortbl}
\usepackage{caption}

\usepackage[pagebackref=true,breaklinks=true,colorlinks,bookmarks=false]{hyperref}

\def\cvprPaperID{4853} % *** Enter the CVPR Paper ID here
\def\confYear{CVPR 2021}

\begin{document}

%%%%%%%%% TITLE
\title{Improving Weakly Supervised Visual Grounding \\by Contrastive Knowledge Distillation}

\author{Liwei Wang $^1$ \!\!\thanks{equal contribution. Contact: lwwang@cse.cuhk.edu.hk} \quad Jing Huang $^{2*}$ \quad 
Yin Li $^3$ \quad Kun Xu $^4$
\quad Zhengyuan Yang $^5$ 
\quad Dong Yu $^4$ \\
$^1$ The Chinese University of Hong Kong \quad
$^2$ University of Illinois at Urbana-Champaign
\\
$^3$ 
University of Wisconsin-Madison
\quad
$^4$
Tencent AI Lab, Bellevue
\quad 
$^5$
University of Rochester

}

\maketitle
\pagestyle{empty}
\thispagestyle{empty}

%%%%%%%%% ABSTRACT
\begin{abstract}
Weakly supervised phrase grounding aims at learning region-phrase correspondences using only image-sentence pairs. A major challenge thus lies in the missing links between image regions and sentence phrases during training. To address this challenge, we leverage a generic object detector at training time, and propose a contrastive learning framework that accounts for both region-phrase and image-sentence matching. Our core innovation is the learning of a region-phrase score function, based on which an image-sentence score function is further constructed. Importantly, our region-phrase score function is learned by distilling from soft matching scores between the detected object names and candidate phrases within an image-sentence pair, while the image-sentence score function is supervised by ground-truth image-sentence pairs. The design of such score functions removes the need of object detection at test time, thereby significantly reducing the inference cost. Without bells and whistles, our approach achieves state-of-the-art results on visual phrase grounding, surpassing previous methods that require expensive object detectors at test time.
\end{abstract}

%%%%%%%%% BODY TEXT

\section{Introduction}
\label{intro}
\begin{figure}
\centering
\includegraphics[width=1\columnwidth]{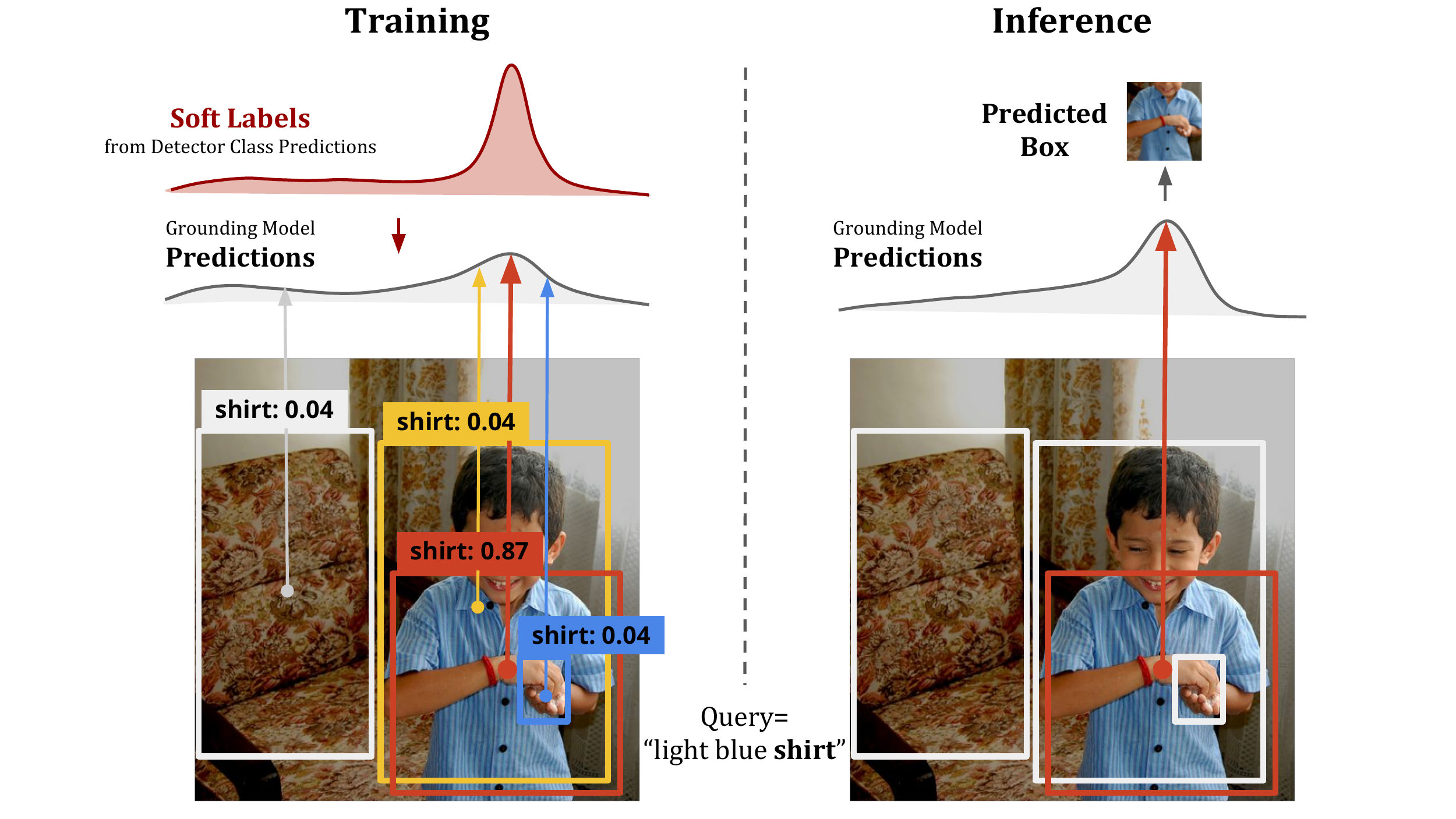}\vspace{-0.5em}
\caption{Our method uses object detector predictions to guide the learning of region-phrase matching in training. At the inference time, our method no longer requires object detectors and directly predicts the box with the highest score.}
\label{fig:teaser}\vspace{-1.5em}
\end{figure}

Visual phrase grounding --- finding regions associated with phrases in a sentence description of the image, is an important problem at the intersection of computer vision and natural language processing. Most of the existing approaches~\cite{fukui2016multimodal,plummer2015flickr30k,wang2018learning} follow a fully supervised paradigm that requires the labeling of bounding boxes for each phrase. These fine-grained annotations are unfortunately expensive to obtain and thus difficult to scale. Consequently, weakly supervised grounding has recently received considerable attention~\cite{rohrbach2016grounding,yeh2017interpretable,yeh2018unsupervised,zhao2018weakly,chen2018knowledge,zhang2018grounding,zhang2018top,fang2019modularized,wang2019phrase}. In this setting, only images and their sentence descriptions are given at training time. At inference time, given an image sentence pair, a method is asked to link regions to sentence phrases.

A major challenge of weakly supervised grounding is to distinguish among many ``concurrent'' visual concepts. For example, the region of a dog and that of its head are likely to co-occur in images associated with the phrase ``a running puppy.'' Without knowing the ground-truth region-phrase matching, learning to link the region of dog (but not dog head) to its corresponding phrase becomes very challenging. To address this challenge, recent methods~\cite{chen2018knowledge, fang2019modularized, wang2019phrase, gupta2020contrastive} leverage generic object detectors for training and/or inference. A detector provides high quality object regions, as well as their category labels that can be further matched to candidate phrases, thereby bringing in external knowledge about region-phrase matching and thus helping to disambiguate those ``concurrent'' concepts. However, it remains unclear about the best practices of using an external object detector for weakly supervised grounding.

In this paper, we focus on developing a principled approach to distill knowledge from a generic object detector for weakly supervised phrase grounding. To this end, we present a simple method under the framework of contrastive learning. Specifically, our model learns a score function between region-phrase pairs, guided by two levels of similarity constraints encoded using noise-contrastive estimation (NCE) loss~\cite{gutmann2010noise} during training. The first level of region-phrase similarity is distilled from object detection outputs. This is done by aligning predicted region-phrase scores to a set of soft targets, computed by matching object names and candidate phrases. The second level of image-sentence similarity is computed from a greedy matching between all region-phrase pairs, and supervised by ground-truth image-sentence pairs. During inference, our method compares each image region to candidate phrases using the learned score function, without the need of object detection. Our training and inference stages are shown in Fig.\ \ref{fig:teaser}.

To evaluate our method, we conduct extensive experiments on Flickr30K Entities~\cite{plummer2015flickr30k} and ReferItGame~\cite{kazemzadeh2014referitgame} datasets. We compare our results to the latest methods of weakly supervised phrase grounding. Our experiments show that our method establishes new state-of-the-art results and outperforms all previous methods, including those using strong object detectors at test time~\cite{fang2019modularized,wang2018learning,yeh2018unsupervised} or using a similar contrastive loss~\cite{gupta2020contrastive}. For example, on Flickr30K Entities, without additional training data, our method outperforms the best reported results by a large margin. On ReferItGame, our method significantly
beats the best reported results. Moreover, we systematically vary the components of our model and demonstrate several best practices for weakly supervised phrase grounding. We hope that our simple yet strong method will shed light on new ideas and practices for weakly supervised image-text grounding.

% Our method also remains very efficient during inference without using object detectors. Finally, we systematically vary the components of our model and demonstrate several best practices for weakly supervised phrase grounding. We hope that our simple yet strong method will shed light on new ideas and practices for weakly supervised image-text grounding.

% we systematically vary the components of our model and demonstrate several best practices for weakly supervised phrase grounding. Moreover, we conduct extensive experiments on Flickr30K Entities~\cite{plummer2015flickr30k} and ReferItGame~\cite{kazemzadeh2014referitgame} datasets, and compare our results to the latest methods of weakly supervised phrase grounding. Our experiments show that our method establishes new state-of-the-art results and outperforms all previous methods, including those that use strong object detectors at test time~\cite{fang2019modularized,wang2018learning,yeh2018unsupervised}, and remains very efficient during inference without using object detectors. Our method also outperforms a latest method that also considered a contrastive loss~\cite{gupta2020contrastive}. For example, on Flickr30K Entities, our method outperforms state of the art by at least 2.6\% in top-1 recall. We hope that our simple yet strong method will shed light on new ideas and practices for weakly supervised image-text grounding.

\section{Related Work}
\label{rela_work}
We discuss relevant works on weakly supervised phrase grounding, and provide a brief review of recent works on contrastive learning and knowledge distillation --- the two main pillars of our method. 

\noindent \textbf{Visual Grounding of Phrases}.
Grounding of textual phrases, also referred to as phrase localization, is an important problem in vision and language. Several datasets, e.g., Flickr30K Entities~\cite{plummer2015flickr30k}, ReferItGame~\cite{kazemzadeh2014referitgame} and Visual Genome~\cite{krishna2017visual}, have been constructed to capture dense phrase-region correspondences. Building on these datasets, many recent approaches learn a similarity function between regions and phrases by using the ground-truth region-phrase pairs~\cite{wang2018learning,fukui2016multimodal,Yang_2019_ICCV,dogan2019neural,plummer2018conditional,Bajaj2019g3,yang2020improving}. More recent works considered phrase grounding in videos~\cite{zhou2019grounded}. This fully supervised paradigm has shown impressive results, yet requires labor-intensive annotations of bounding boxes for all phrases.

\noindent \textbf{Weakly Supervised Phrase Grounding}.
Weakly supervised phrase grounding has thus received considerable attention recently~\cite{rohrbach2016grounding,yeh2018unsupervised,xiao2017weakly,chen2018knowledge, zhao2018weakly,fang2019modularized,wang2019phrase,gupta2016cross}. In this setting, a method learns from only images and paired sentence descriptions, without explicit region-to-phrase correspondence. 

Recent works~\cite{chen2018knowledge, yeh2018unsupervised,yeh2017interpretable} show that weakly supervised phrase localization can benefit from side information, such as object segmentation or detection. For example, Chen et al.\ \cite{chen2018knowledge} leveraged pre-trained deep models and proposed to enforce visual and language consistency. Yeh et al.\ \cite{yeh2018unsupervised} proposed to link words in text and detection classes using co-occurrence statistics from paired captions. Moreover, Xiao et al.\ \cite{xiao2017weakly} investigated the linguistic structure of the sentences. They proposed a structure loss to model the compositionality of the phrases and their attention masks. Zhao et al.\ \cite{zhao2018weakly} presented a model that jointly learns to propose object regions and matches the regions to phrases. Fang et al.~\cite{fang2019modularized} explored the weakly supervised grounding by decomposing the problem into several modules and taking additional information, like the color module, to improve the performance. More recently, Wang et al.\  \cite{wang2019phrase} made use of off-the-shelf models to detect objects, scenes and colors in images, and achieves the goal of grounding via measuring semantic similarity between the categories of detected visual elements and the sentence phrases. Gupta et al.\ \cite{gupta2020contrastive} proposed to use contrastive loss for weakly supervised phrase grounding. Datta et al.\ \cite{datta2019align2ground} adopted image-sentence retrieval to guide the phrase localization.

Our method shares the key idea behind these previous works--we seek an explicit alignment between regions and phrases given image-sentence pairs. Our method differs from previous works by explicitly modeling the knowledge distillation from external off-the-shelf object detectors into a unified contrastive learning framework. The most relevant work to us is InfoGround~\cite{gupta2020contrastive}. Both our method and InfoGround~\cite{gupta2020contrastive} consider contrastive learning for weakly supervised grounding. Different from~\cite{gupta2020contrastive}, our work moves beyond the contrastive loss and focuses on knowledge distillation from an external object detector under a contrastive learning framework. Our method is also different from WPT~\cite{wang2019phrase} though we both need to use object detectors: our approach learns the distillation in the training stage which makes our model free of detectors in the testing stage, while WPT requires using detectors during inference.

\noindent \textbf{Contrastive Learning}.
There has been a recent trend of exploring contrastive loss for representation learning. For instance, Oord et al.\ \cite{oord2018representation} proposed Contrastive Predictive Coding (CPC) that learns representations for sequential data. Hjelm et al.\ \cite{hjelm2018learning} presented Deep InfoMax for unsupervised representation learning by
maximizing the mutual information between the input and output of a deep network. More recently, Chen et al.\ \cite{chen2020simple} proposed to learn visual representations by maximizing agreement between differently augmented
views of the same image via a contrastive loss. He et al.\ \cite{he2019momentum} proposed Momentum Contrast (MoCo) for unsupervised visual representation learning. Tian et al.\ \cite{tian2019contrastive} extends the input to more than two views. These methods are all based on a similar contrastive loss related to Noise Contrastive Estimation (NCE)~\cite{gutmann2010noise}. The NCE loss has also been explored for phrase grounding by Gupta et al. \cite{gupta2020contrastive} (InfoGround). Our work is relevant to these works since our mathematical framework is also built upon the general idea of infoNCE~\cite{oord2018representation} and NCE~\cite{gutmann2010noise}. However, our work leverages a contrastive loss for knowledge distillation in the context of cross-view weakly supervised grounding.

\noindent \textbf{Knowledge Distillation}.
Knowledge distillation was proposed and popularized by~\cite{bucilu2006model,hinton2015distilling,ba2014deep,romero2014fitnets,zagoruyko2016paying}. Several recent works~\cite{gupta2016cross,garcia2018modality,luo2018graph,li2017large,tian2019contrastive} explored knowledge distillation for multi-modal learning. Knowledge distillation has also shown its effectiveness in various vision-language tasks, such as VQA~\cite{mun2018learning,do2019compact}, grounded image captioning~\cite{zhou2020more} and video captioning~\cite{pan2020spatio,zhang2020object}. Different from previous approaches, we consider knowledge distillation for region-phrase grounding by matching the outputs of a region-phrase score function to soft targets computed from object detection results.

\section{Approach}
\label{approach}
\begin{figure*}[t]
\centering
\includegraphics[width=0.7\textwidth]{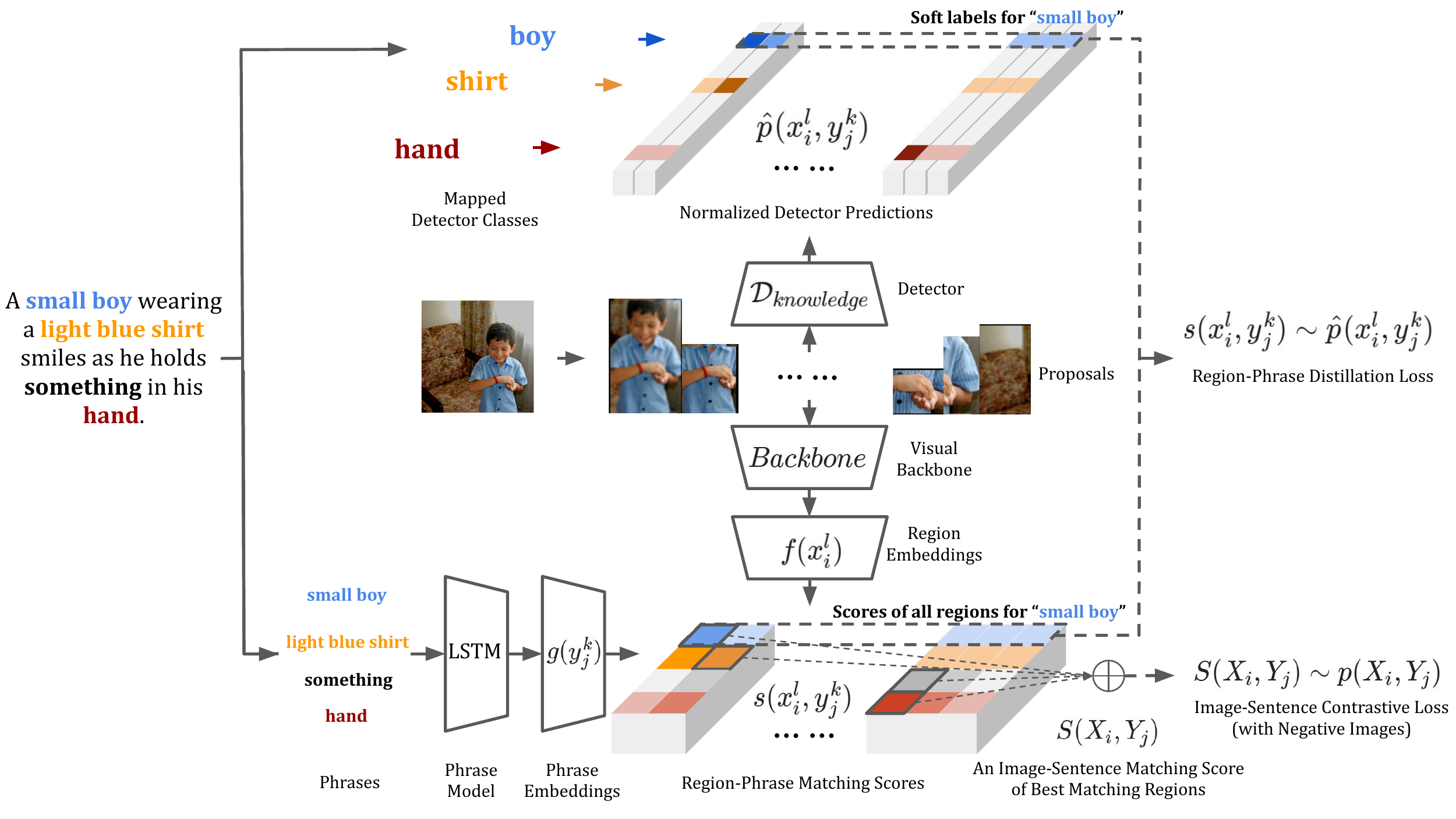}\vspace{-0.5em}
\caption{Overview of our method in training. A contrastive learning framework is designed to account for both region-phrase and image-sentence matching. The top part illustrates region-phrase matching learned by distilling from object detection outputs, while the bottom part shows image-sentence matching supervised by ground-truth image-sentence pairs.}\label{fig:architecture}\vspace{-1.0em}
\end{figure*}

Consider $X = [X_1, ..., X_i, ..., X_N]$ as the set of images and $Y = [Y_1, ..., Y_j, ..., Y_M]$ as the set of sentences. Each image $i$ consists of a set of regions with their features $X_i = [x_i^1, ..., x_i^l, ..., x_i^n]$. Similarly, each sentence $j$ includes multiple phrase features $Y_j = [y_j^1, ..., y_j^k, ..., y_j^m]$. Thus, $i, j$ index images and sentences, and $l, k$ index regions and phrases. Oftentimes, we have multiple sentences describing the same image and many more image regions than sentence phrases. Moreover, with minor abuse of notations, we denote $p(X_i, Y_j)$ as the probability of a valid image-sentence pair $(X_i, Y_j)$, i.e., $p(X_i, Y_j) = 1$ if and only if $Y_j$ can be used to describe $X_i$. Similarly, we use $p(x_i^l, y_j^k)$ as the probability of a valid region-phrase pair $(x_i^l, y_j^k)$.

Our goal is to learn a score function that measures the similarity between region features $x_i^l$ and phrase features $y_j^k$. However, we only have access to ground-truth image-sentence pairs $p(X_i, Y_j)$ without knowing the matching between regions and phrases $p(x_i^l, y_j^k)$. To address this challenge of \emph{weakly supervised} grounding, we leverage a generic object detector $\mathcal{D}$ to label candidate image regions, then generate ``pseudo'' labels of region-phrase correspondence by matching the region object labels to the sentence phrases. Therefore, our key innovation is the design of a contrastive loss that learns to distill from object detection outputs. A main advantage of using knowledge distillation is that our method no longer requires object detection at inference time and thus is very efficient during inference. 

Fig.\ \ref{fig:architecture} presents an overview of our method. We now present the details of our method by first introducing the design of our score functions for image-text matching, followed by our contrastive learning loss using knowledge distillation. 

\subsection{Score Functions for Image-Text Matching}

Our model builds on a two-branch network~\cite{wang2018learning} for image-text matching at both region-phrase and image-sentence levels. The key idea is learning a score function to match region-phrase pairs. Based on the region-phrase matching scores, we further construct an image-sentence similarity score. Specifically, our network has two branches $f$ and $g$ that take the inputs of region and phrase features $x_i^l$ and $y_j^k$, respectively. Each branch is realized by a deep network by stacking multiple fully connected layers with ReLU activation in-between, followed by a L$2$ normalization at the end. We define the similarity between a region-phrase pair $(x_i^l, y_j^k)$ as the cosine similarity between the transformed features $f(x_i^l)$ and $g(y_j^k)$, given by
\begin{equation} \label{eq:noweight}
\small
s(x_i^l, y_j^k) = f(x_i^l)^T g(y_j^k).
\end{equation}

We further aggregate the region-phrase matching scores $s(x_i^l, y_j^k)$ into a similarity score between a image-sentence pair $(X_i, Y_j)$, defined as
\begin{equation} \label{eq:sim_score}
\small
S(X_i, Y_j) = \sum_{k=1}^{m} \max_{{1\leq l\leq n}} \ s(x_i^l, y_j^k).
\end{equation}
This image-sentence score $S(X_i, Y_j)$ is computed using greedy matching. Concretely, for each phrase $k$ in the sentence $j$, we find its best matching region in an image.
%\textcolor{red}{This objective naturally corresponds to the “max” operation, as at least one region in the positive image should have a high score, while all regions in the negative images should have low scores.}
The scores of best matching regions are further summed across all phrases. Note that phrases and regions are not interchangeable in this score function, i.e., $S(X_i, Y_j)\neq S(Y_j, X_i)$, because each phrase must be matched to at least one region, while some regions, such as background regions, are not matched to any phrase. Similar image-sentence scoring functions were discussed in~\cite{karpathy2014deep,zhou2018weakly} for image-sentence retrieval.

\subsection{Distillation using Contrastive Learning}

A major challenge of weakly supervised grounding is the lack of ground-truth region-phrase pairs. Our key idea is to make use of an object detector during training that can provide ``pseudo'' labels. Our model further distills from these ``pseudo'' labels for learning region-phrase matching. Once learned, we can directly use the region-phrase score function without object detection. In what follows, we describe the generation of pseudo labels, the contrastive loss used for distillation, and the training and inference of our model. 

\noindent \textbf{Pseudo Labels for Region-Phrase Matching}. An object detector $\mathcal{D}$ predicts the probability of region $x_i^l$ having an object label $z_i^l$ in the form of nouns (including ``background''), i.e., $p(z_i^l|x_i^l)=\mathcal{D}(x_i^l)$.
$z_i^l$ can be further matched to the phrase $y_j^k$, e.g., using similarity scores between object noun and the head noun of the phrase. {In particular, our implementation uses the WordNet~\cite{miller1995wordnet} to define such similarity scores, as discussed in section~\ref{exp}.} Let $p(y_j^k, z_i^l)$ be the matching probability between $y_j^k$ and $z_i^l$. We propose to approximate the unknown region-phrase matching ground-truth $p(x_i^l, y_j^k)$ by soft ``pseudo'' label $\hat{p}(x_i^l, y_j^k)$, approximated by

\begin{equation}
\label{eq:eq_3}
\small
    \hat{p}(x_i^l, y_j^k) \propto \sum_z p(y_j^k, z_i^l) p(z_i^l|x_i^l) p(x_i^l).
\end{equation}
where we assume $p(z_i^l)$ is a constant value for every $z_i^l$ since object classes are fixed and limited and therefore we omit the $p(z_i^l)$ in the denominator of Eq.\ref{eq:eq_3}. Note that this approximation holds when $x_i^l$ and $y_j^k$ is conditionally independent given $z_i^l$ --- a rather strong assumption. Nonetheless, we use $\hat{p}(x_i^l, y_j^k)$ as a ``pseudo'' soft target distribution for training our model. In practice, $\hat{p}(x_i^l, y_j^k)$ is approximated and computed by detecting objects and matching their names to candidate phrases.

\noindent \textbf{Distilling Knowledge from Pseudo Labels}. We propose to distill from the pseudo label $\hat{p}(x_i^l, y_j^k)$ by aligning the region-phrase matching scores $s(x_i^l, y_j^k)$ to the soft pseudo label $\hat{p}(x_i^l, y_j^k)$. Specifically, given a matching image-sentence pair $(X_i, Y_j)$, we propose the following distillation loss function for region-phrase matching
\begin{equation} \label{eq:obj_distill}
\small
\mathcal{L}_{RP}(X_i, Y_j) = 
- \sum_{y_{j}^{k}\in Y_j} \sum_{x_{i}^{l} \in R_i^l} 
\hat{p}(x_i^l, y_j^k) 
\log \hat{h}(x_i^l, y_j^k), 
\end{equation}
where $\hat{h}(x_i^l, y_j^k)$ is given by
\begin{equation*}
\resizebox{.45\textwidth}{!} 
{
    $\hat{h}(x_i^l, y_j^k) =  
    \frac{\exp(s(x_{i}^{l},y_{j}^{k})/\tau)}{\exp(s(x_{i}^{l},y_{j}^{k})/\tau) + \sum_{x_{i}^{l'} \in {R_i^l} {\setminus \{x_i^l\}}} \exp(s(x_{i}^{l'},y_{j}^{k})/\tau)}$.
}
\end{equation*}
Here $\tau$ is the temperature scale factor (0.5 in all our experiments). $R_i^l$ controls how we select $x_{i}^{l'}$. {A simple choice for $x_{i}^{l'}$ is to use all regions in $X_i$ except $x_{i}^{l}$.} In this case, our loss can be interpreted as the cross entropy loss, where the normalized output of the score function $s(x_{i}^{l},y_{j}^{k})$ is trained to mimic the pseudo label $\hat{p}(x_i^l, y_j^k)$ given by object detection outputs. This is the same idea as knowledge distillation~\cite{hinton2015distilling}, where the soft target from a teacher detection model is used as a learning objective.

\noindent \textbf{Contrastive Loss for Image Sentence Matching}. Moving beyond region-phrase matching, we enforce additional constraints for image-sentence matching scores $S(X_i, Y_j)$, where the ground truth pairs $p(X_i, Y_j)$ is readily available.
%\textcolor{red}{The image-sentence level matching helps to bridge the gap between the predefined detector labels and free-form natural language queries.}
To this end, we make use of the noise contrastive estimation loss~\cite{gutmann2010noise} to contrast samples from data distribution (matched pairs) and noise distribution (non-matched pairs). The NCE loss for image-sentence matching is thus given by
\begin{equation} \label{eq:obj_contrast}
\small
\mathcal{L}_{IS}(X_i, Y_j) = -\mathbb{E}_{\mathcal{N}(X_i) \in X} \left[ 
p(X_i, Y_j) \\
\log h(X_i, Y_j)
\right],
\end{equation}
where $h(X_i, Y_j)$ is defined as
\begin{equation*}
\resizebox{.45\textwidth}{!} 
{
    $h(X_i, Y_j) = \frac{\exp(S(X_i,Y_j)/\tau)}{\exp(S(X_i,Y_j)/\tau) + \sum_{i' \in \mathcal{N}(X_i)} \exp(S(X_{i'},Y_j)/\tau)}$.
}
\end{equation*}
Again, $\tau$ is the temperature scale factor (0.5). $p(X_i, Y_j)$ is reduced to binary values during training, i.e., $p(X_i, Y_j)=1$ if and only if $(X_i, Y_j)$ is a ground-truth image-sentence pair. $\mathcal{N}(X_i)$ includes a set of negative samples, i.e., those images not matched to the current sentence $Y_j$, sampled from the set of images $X$. In practice, we always sample a fixed number of negative pairs from the current mini-batch. 

\noindent \textbf{Loss Function}. We note that both Eq.\ \ref{eq:obj_distill} and Eq.\ \ref{eq:obj_contrast} share a similar form and can be both considered as a variant of contrastive loss. Concretely, the two loss functions seek to align the normalized scores in the form of NCE to a target distribution. The difference is how the target distribution is defined and how the samples are selected for normalization. For region-phrase matching, the target distribution is given by pseudo labels from object detection and local image regions are used for normalization. For image-sentence matching, the target distribution is defined by ground-truth image-sentence pairs and non-matched image-sentence pairs are sampled for normalization. 

By combining the distillation loss  $\mathcal{L}_{RP}(X_i, Y_j)$ for region-phrase matching and the NCE loss $\mathcal{L}_{IS}(X_i, Y_j)$ for image-sentence matching, our final loss function is given by  
\begin{equation} \label{eq:obj_total}
\small
\mathcal{L}(X_i, Y_j) = \mathcal{L}_{IS}(X_i, Y_j) + \lambda\mathcal{L}_{RP}(X_i, Y_j),
\end{equation}
where $\lambda$ is the coefficient balancing the two loss terms. During training, we gradually increase the coefficient $\lambda$, such that our model learns to optimize image-sentence matching during the early stage of training, and to focus on region-phrase matching during the late stage of training.

\noindent \textbf{Inference without Object Detection}. During inference, given an input image-sentence pair, we apply the learned region-phrase score function $s(x_{i}^{l},y_{j}^{k})$ between every region-phrase pair. The image region with the highest score to each phrase is then selected as the grounding results, i.e., $\arg \max_{x_{i}^{l}} s(x_{i}^{l},y_{j}^{k})$. We must point out that unlike previous methods~\cite{wang2019phrase,gupta2020contrastive} {\it the inference of our model does not require running object detection}, therefore our method is very efficient at test time.

\section{Experiments and Results}
\label{exp}
We now present our experiments and results. We first discuss our datasets, experiment setup and implementation details, followed by a comparison of our results to latest methods and an ablation study of our model.

\noindent \textbf{Datasets}. Our experiments are conducted on two major visual grounding datasets: Flickr30K Entities~\cite{plummer2015flickr30k} and the ReferItGame dataset~\cite{kazemzadeh2014referitgame}. Flickr30K Entities~\cite{plummer2015flickr30k} includes around 30K images. Each image is associated with five sentences. We follow the same train/val/test splits from~\cite{plummer2015flickr30k}. ReferItGame~\cite{kazemzadeh2014referitgame} includes 20K images and 120K phrases. For ReferItGame, we follow the standard split of~\cite{rohrbach2016grounding}.

\noindent \textbf{Experiment Setup}. We follow the setting of weakly supervised grounding, and do not use the region-phrase annotations of both datasets during training. For evaluation, we follow the standard protocol used in~\cite{chen2018knowledge,rohrbach2016grounding,wang2019phrase,gupta2020contrastive}, we report accuracy as the evaluation metric. Accuracy is defined as the fraction of query phrases whose predicted bounding box overlaps ground-truth box with $IoU$>$0.5$. For methods that select the predicted bounding box from a set of region proposals, the reported accuracy metric is equivalent to $Recall@1$, as used in \cite{gupta2020contrastive}.

\subsection{Implementation Details}
We first describe our implementation details, including the features and object detectors, the network architecture and training scheme, and details of object-phrase matching. %Our code will be made publicly available.

\noindent \textbf{Features and Object Detectors}.
To establish a fair comparison with previous work using region features extracted from different backbones, we benchmark our methods by varying the backbone networks. We follow the same settings in~\cite{chen2018knowledge,wang2019phrase} to extract activations from the last layer before the classification head in Faster R-CNN~\cite{ren2015faster} with VGG16 and ResNet-101 backbones pre-trained on PASCAL VOC (PV)~\cite{everingham2009pascal} or MS COCO (CC)\footnote{\url{https://github.com/endernewton/tf-faster-rcnn}}~\cite{lin2014microsoft}. To compare with WPT~\cite{wang2019phrase} using object detectors trained on Open Images Dataset~\cite{krasin2017openimages}, we  also extract classifier logits from Faster R-CNN with Inception-ResNet-V2 (IRV2) backbone pre-trained on the Open Images Dataset (OI)\footnote{\url{https://github.com/tensorflow/models/blob/master/research/object_detection/g3doc/detection_model_zoo.md}}. To compare with the InfoGround from~\cite{gupta2020contrastive}, we also experiment with their released features\footnote{\url{https://github.com/BigRedT/info-ground}} on Flickr30K and follow their protocols to use the same VisualGenome (VG) pre-trained Faster R-CNN model\footnote{\url{https://github.com/BigRedT/bottom-up-features}} to extract features on ReferItGame. Unlike  \cite{chen2018knowledge,rohrbach2016grounding,zhao2018weakly} using a large amount of bounding box proposals, typically $100$ per image, InfoGround only extracts $30$ proposals per image for the grounding model to select from. We compare the two settings separately in the result tables.

We denote these feature choices as ``{\it VGG16}'', ``{\it Res101}'', ``{\it IRV2}'' respectively plus the object data set when reporting our results. For example, ``{\it IRV2 OI}'' means that the backbone is Inception-ResNet-V2 (IRV2) pre-trained on the Open Images (OI) Dataset.

\renewcommand{\arraystretch}{0.7}
\begin{table*}[th!]
  \small
  \centering
  \resizebox{1.8\columnwidth}{!}{%
  \begin{tabular}{lcccc}
    \toprule
   \parbox{3cm}{Method}  &
   Backbone &
   Detector$_{K}$  &
   Require Detector$_{K}$ in Inference &
   ACC (\%) \\ \midrule
GroundeR ~\cite{rohrbach2016grounding} & VGG16 PV & - & - & 28.94 \\
\cmidrule(lr){1-5}
MATN ~\cite{zhao2018weakly} & VGG16 PV & - & - & 33.10 \\
\cmidrule(lr){1-5}

\multirow{2}{*}{UTG~\cite{yeh2018unsupervised}} & - & VGG16 PV & Yes & 35.90 \\
& - &  YOLOv2 CC &  Yes & 36.93  \\
\cmidrule(lr){1-5}

% + Soft KBP
\multirow{2}{*}{\parbox{3.5cm}{KAC~\cite{chen2018knowledge}}} & VGG16 PV &  VGG16 PV &  Yes &36.14  \\
 & VGG16 PV &  VGG16 CC &  Yes & 38.71 \\
\cmidrule(lr){1-5}

\multirow{2}{*}{\parbox{3.5cm}{MTG %Entity+Attr+Col
\cite{fang2019modularized}}} & Res101 CC+Res50 CC &  - &  - & 48.66 \\
 &  +Res50 CL &  &  &  \\
\cmidrule(lr){1-5}

\multirow{4}{*}{\parbox{3.5cm}{WPT~\cite{wang2019phrase}\\(w2v-max union)}} & - &  IRV2 CC &  Yes & 37.57 \\
 & - &  IRV2 CC+IRV2 OI &  Yes & 48.20 \\
 & - &  IRV2 CC+IRV2 OI &  Yes & 50.49 \\
  &  &  +WRN18 PL & & \\
\cmidrule(lr){1-5}

% Align2Ground reports a different metric - the pointing accuracy
% Predict a single point location per phrase and the prediction is counted as correct if it falls within the ground truth bounding box for the phrase
% Align2Ground~\cite{datta2019align2ground} & Res101 VG & - & - & - \\
%\cmidrule(lr){1-5}

InfoGround~\cite{gupta2020contrastive} (Trained on Flickr30K) & Res101 VG &  - &  - & 47.88  \\
\quad\quad\quad\quad\quad\quad\quad(Trained on COCO) 
 & Res101 VG & - & - & 51.67 \\
\cmidrule(lr){1-5}

%NCE (Ours)  & Res101 VG & - & - & \textbf{51.31} \\
%\cmidrule(lr){1-5}

\multirow{3}{*}{\parbox{3.5cm}{NCE+Distillation (Ours)}} & VGG16 PV &  VGG16 CC &  No & 40.38 \\
 & Res101 CC &  IRV2 OI &  No & 50.96 \\
 & Res101 VG &  IRV2 OI &  No & \textbf{53.10} \\
    \bottomrule
  \end{tabular}%
}\vspace{-0.5em}
\caption{Results on Flickr30K Entities. We report phrase localization accuracy and list the settings of different methods. ``Backbone'' denotes the visual backbone used to extract region features. Detector$_{K}$ denotes the detector that provides external knowledge. ``-'' denotes the method does not use backbone or Detector$_{K}$.  Dataset notations: PV=PASCAL VOC, CC=COCO, OI=Open Images, CL=Color Name, PL=Place365, and VG=Visual Genome.} \label{tab:flickr30k-results}\vspace{-1.5em}
\end{table*}

\noindent \textbf{Network Architecture}.
For visual representation, we normalized the region features to zero-mean and unit-variance using stats from training samples. This normalization helps our model to converge faster.
We attached two fully connected layers  on top of the region features to get $512$-D region embeddings.
For phrase representation, we tokenized each query phrase into words
and encoded using LSTM~\cite{hochreiter1997long} with the Glove embeddings~\cite{pennington2014glove}.
The embedding vocabulary contains the most frequent 13K tokens from the Flickr30K Entities training split. The same vocabulary is used for ReferItGame. The LSTM has two layers, with both embedding and hidden dimension as $300$. Max pooling is applied over the hidden states of all tokens, followed by two fully connected layers to get $512$-D phrase embeddings.

\noindent \textbf{Training Details}.
We trained our model using Adam\cite{kingma2014adam} with a learning rate of 0.0001. We used a mini-batch size of 32 image-sentence pairs (31 negative images per sentence for the contrastive loss). Unlike~\cite{fang2019modularized}, we did not fine-tune our vision backbone during training for efficiency. Similarly, the GloVe embeddings~\cite{pennington2014glove} are also fixed during training. We observed that the model converges quickly within a few epochs on both datasets. For the $\lambda$ in Eq. \ref{eq:obj_total}, we gradually increased the value using a staircase function $\lambda=\min(\lfloor step / a \rfloor, b)$, where $a, b$ are selected based on the validation set. We observed that $a$=$200$, $b$=$1$ for VG features and $a$=$200$, $b$=$3$ for others work the best.
%Our implementation is in TensorFlow and will be made publicly available.

\noindent \textbf{Object-Phrase Matching}.
We use the WordNet~\cite{miller1995wordnet} to define the similarity scores between phrases and region object labels. Specifically, we identify the head noun of each phrase using the off-the-shelf POS tagger provided by NLTK~\cite{loper2002nltk}, which uses the Penn Treebank tag set. If the head noun matches one of the detector class labels, the phrase is mapped to the class. Otherwise, we look up the head noun in the WordNet~\cite{miller1995wordnet} to find its corresponding synset, the synset's lemmas, and hypernyms. If any of these exists in the detector classes, the phrase is mapped to the class. For phrases with multiple synsets, the most frequent one is used. The WordNet synset helps to match phrases such as ``spectators'' to ``person'' and ``sweater'' to ``clothing''. With the $545$ classes in Open Images Dataset~\cite{krasin2017openimages}, the WordNet-based matching algorithm covers $18$k out of $70$k unique phrases in Flickr30k Entities and $7$k out of $27$k in ReferItGame training set.

We empirically observe that using WordNet is more reliable than word embeddings for noun matching. Similarity in the word embedding space is not necessarily aligned with the visual similarity of the entities described. A similar observation was made in~\cite{fang2019modularized} where the word2vec embeddings are not discriminative for gender related visual concepts.

\subsection{Comparison to Other Methods}
We further compare our results to the latest methods of weakly supervised phrase grounding on both Flickr30K Entities and ReferItGame datasets.

\noindent \textbf{Baselines} We consider a number of baselines. Our main competitors are those methods using object detectors, including KAC~\cite{chen2018knowledge}, UTG~\cite{yeh2018unsupervised}, and WPT~\cite{wang2019phrase}.
Among these methods, KAC and UTG used detectors during both training and inference. WPT applied detectors during inference. While these baselines have very different sets of detectors and backbones, we try to match their settings in our experiments. Our baselines also include previous methods that do not use object detectors, such as GroundeR~\cite{rohrbach2016grounding}, MATN~\cite{zhao2018weakly}, MTG~\cite{fang2019modularized}, and InfoGround~\cite{gupta2020contrastive} for completeness.

\renewcommand{\arraystretch}{0.7}
\begin{table*}[th!]
  \small
  \centering
  \resizebox{1.8\columnwidth}{!}{%
  \begin{tabular}{lcccc}
    \toprule
  \parbox{3cm}{Method}  &
  Backbone & 
  Detector$_{K}$  &
  Require Detector$_{K}$ in Inference &
  ACC (\%) \\
    \midrule
GroundeR ~\cite{rohrbach2016grounding} & VGG16 PV & - & - & 10.70 \\
\cmidrule(lr){1-5}
MATN ~\cite{zhao2018weakly} & VGG16 PV & - & - &  13.61 \\
\cmidrule(lr){1-5}

UTG ~\cite{yeh2018unsupervised} & - & VGG16 CC+YOLOv2 CC & Yes & 20.91  \\
\cmidrule(lr){1-5}

%  + Soft KBP
\multirow{2}{*}{\parbox{3cm}{KAC~\cite{chen2018knowledge}}} & VGG16 PV & VGG16 PV & Yes & 13.38  \\
 & VGG16 PV & VGG16 CC & Yes & 15.83 \\
\cmidrule(lr){1-5}

\multirow{4}{*}{\parbox{3.5cm}{WPT~\cite{wang2019phrase} \\(w2v-max union)}} & - & IRV2 CC & Yes & 15.40 \\
 & - & IRV2 CC+IRV2 OI & Yes & 26.48 \\
  &  & +WRN18 PL+CL & & \\
\cmidrule(lr){1-5}

\multirow{3}{*}{\parbox{3cm}{NCE+Distillation (Ours)}} & VGG16 PV & VGG16 CC & No & 24.52 \\
 & Res101 CC & IRV2 OI &  No & 27.59 \\
  & Res101 VG & IRV2 OI & No & \textbf{38.39} \\
    \bottomrule
  \end{tabular}%
}\vspace{-0.5em}
  \caption{Results on ReferItGame. We report phrase localization accuracy and settings of different methods. ``Backbone'' denotes the visual backbone used to extract region features. Detector$_{K}$ denotes the detector that provides external knowledge. ``-'' denotes the method does not use backbone or Detector$_{K}$. Dataset notations: PV=PASCAL VOC, CC=COCO, OI=Open Images, CL=Color Name, and PL=Place365.}\label{tab:referit-results}\vspace{-1.5em}
\end{table*}

\noindent \textbf{Results on Flickr30K Entities}. Our results on Flickr30K Entities are summarized in Table~\ref{tab:flickr30k-results}. Table~\ref{tab:flickr30k-results} compares both the settings of different methods and their phrase localization accuracy.
When using detectors pre-trained on COCO (YOLOv2 CC, VGG16 CC, IRV2 CC), our method outperforms previous works (+3.5\% / +1.7\% / +2.81\% for UTG / KAC / WPT, respectively). In comparison to MTG which uses COCO pre-trained backbones, our results are better (+2.6\%) with only a single VGG16 backbone in inference (vs.\ three ResNet backbones in MTG).

When using a stronger backbone (Res101 CC) and a better detector pre-trained on a larger scale dataset (IRV2 OI), our results are further improved by 10.6\% on Flickr30K Entities. Our final results thus outperform the latest method of WPT under a similar training setting. Moreover, in contrast to WPT, our method does not require running the three cumbersome object detectors during inference, thus is more applicable for real world deployment. 

Lastly, to compare fairly with~\cite{gupta2020contrastive}, using the same VisualGenome pre-trained backbone (Res101 VG) and under the same proposals setting (30 per image) with~\cite{gupta2020contrastive}, our NCE+Distill model significantly outperforms the latest method of InfoGround by 5.22\% when trained on the Flickr30K Entities dataset. Moreover, with the help of distillation, our results also outperform their best results by 1.4\%, which is trained on COCO Caption dataset~\cite{chen2015microsoft} using a strong language model (BERT~\cite{devlin2018bert}).

\begin{table}[t]
  \centering
  \begin{adjustbox}{width=1\columnwidth}
  \begin{tabular}{lccc}
    \toprule
    Method     &  Detector$_{K}$ GFLOPs & Backbone GFLOPs & Total GFLOPs    \\
    \midrule
   % YOLOv2 stats come from https://pjreddie.com/darknet/yolov2/
    \multicolumn{4}{l}{Detector$_{K}$=YOLOv2 CC}  \\
    UTG  & 60 & - & 60 \\
    \cmidrule(lr){1-4}
    \multicolumn{4}{l}{Detector$_{K}$=Faster R-CNN VGG16 CC} \\
    KAC  & 180 & 180 (VGG16 PV) & 360 \\ 
    NCE+Distill & 0 (Not needed) & 180 (VGG16 PV) & 180 \\
    \cmidrule(lr){1-4}
    \multicolumn{4}{l}{Detector$_{K}$=Faster R-CNN IRV2 OI} \\
    WPT & 1600+ (+IRV2 CC,WRN18 PL) & - & 1600+ \\
    NCE+Distill & 0 (Not needed) & 500 (Res101 CC) & 500 \\ 
    \bottomrule
  \end{tabular}
  \end{adjustbox}\vspace{-0.5em}
  \caption{Estimated number of FLOPs in inference.}\label{tab:flops-results}\vspace{-1.5em}
\end{table}

To further illustrate the benefits of our model, we compare the computation complexity at inference time in terms of floating point operations (FLOPs). For each model, the estimation is a combination of the backbone and the object detector, since the rest part of the model, including the language feature extractor, is computationally insignificant (0.3 GFLOPs for our NCE+Distill model). For Faster R-CNN based detectors, we use the numbers reported in \cite{huang2017detspeed}, under the high-resolution input (600$\times$600) and 300 proposals setting. For YOLOv2, we use the number reported in \cite{redmon2017yolo9000}. The comparison focuses on methods that incorporate external knowledge, namely UTG, KAC, WPT, and ours. As shown in Table \ref{tab:flops-results}, our proposed NCE+Distill method reduces the computational complexity by $50\%$ and $70\%$, while being a bit more expensive than UTG due to the detector meta-architecture difference (Faster R-CNN vs. YOLOv2).

\noindent \textbf{Results on ReferItGame}
We summarize the results on ReferItGame under different settings in Table~\ref{tab:referit-results}. When using COCO pre-trained detectors, our method significantly outperforms UTG, KAC, and WPT by 3.6\%, 8.7\%, and 9.12\% respectively. When using a stronger backbone (Res101 CC) and a better detector pre-trained on a larger scale dataset (IRV2 OI), our results are improved by 3.1\%, outperforming the best results from WPT using four cumbersome knowledge detectors in inference by 1.1\%.

Finally, using the VisualGenome pre-trained backbone (Res101 VG), our results can be further improved by 10.8\%. This gain is significantly larger than the one on Flickr30K Entities. We conjecture that the VisualGenome pre-trained backbone provides more discriminative features in certain categories that appear more frequently in the ReferItGame than Flickr30K Entities. To verify this hypothesis, we compare the most frequent phrases from Flickr30K Entities and ReferItGame. In Flickr30K Entities, top phrases are mostly people related: \textit{man, woman, boy, girl, person, people, dog, two men, street, young boy, child}, whereas top phrases in ReferItGame are mostly scene related: 
\textit{sky, water, people, ground, person, trees,
building, face, road, grass, clouds}. Many scene related objects are not in the COCO label set, but are available in the VisualGenome categories.
%The largely improved performance brought by pre-training also suggests that our proposed contrastive model can effectively mine the correspondences between regions and phrase.

\begin{figure*}[h!]
  \centering
  \includegraphics[clip, trim=0cm 3cm 0cm 0cm,width=0.95\textwidth]{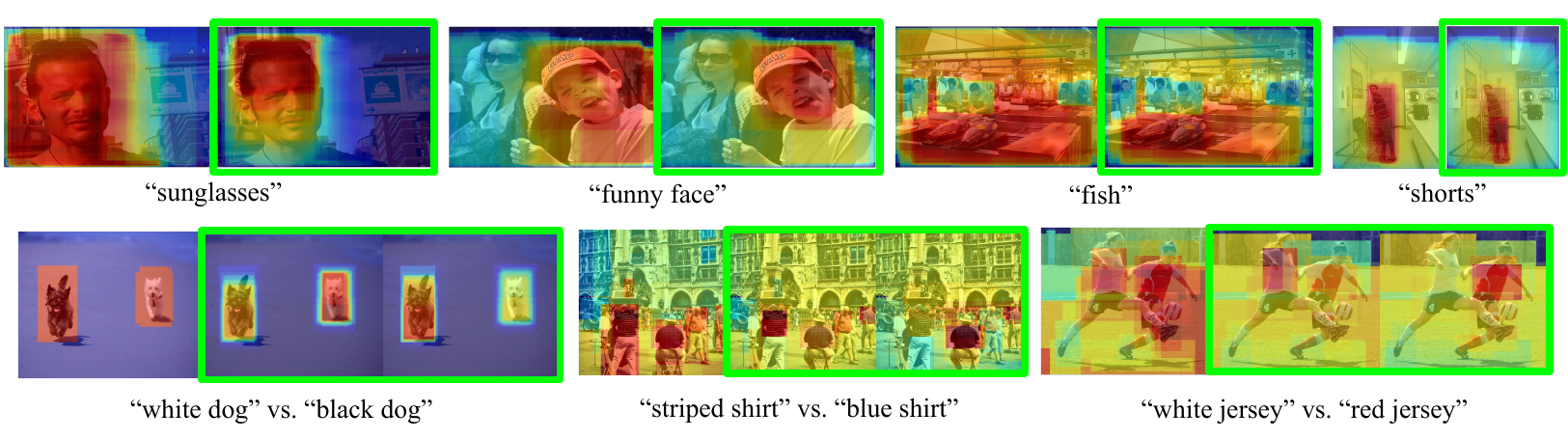}\vspace{-1em}
  \caption{Visualization of region-phrase matching. First row: results of the NCE only (left) and the NCE+Distill (green) on phrases mapped to Open Images Dataset classes. Second row: results of the Distill only (left) and the NCE+Distill (green) on phrases mapped to the same class but with different attributes. For each pixel, we compute a matching score by averaging scores from all proposals covering the pixel. Red colors indicate high matching scores.
  Our knowledge distillation helps to better identify the extent of objects, while contrastive learning helps to distinguish finer attributes.}\label{fig:compare}\vspace{-1.5em}
\end{figure*}

\subsection{Ablation Study}

To fully understand our model, we conduct ablation studies on both Flickr30K Entities and ReferItGame datasets. Specifically, we consider four different variants of our model: (1) our model with only image-sentence score function 
(Eq.\ \ref{eq:sim_score}) supervised by a max margin loss following ~\cite{karpathy2014deep,zhao2018weakly}, denoted as ``Max Margin'', i.e. modeling positive and negative image-sentence pairs distances via max margin loss. The full loss function is defined in the Eq.\ref{eq:margin} below.
(2) our model with only image-sentence score function (Eq.\ \ref{eq:sim_score}) supervised by the NCE loss (Eq.\ \ref{eq:obj_contrast}), denoted as ``NCE''; (3) our model with only region-phrase score function (Eq.\ \ref{eq:noweight}) supervised by the distillation loss (Eq.\ \ref{eq:obj_distill}), denoted as ``Distill''; and (4) our full model with both region-phrase and image-sentence score functions supervised by our joint loss (Eq.\ \ref{eq:obj_total}), denoted as ``NCE+Distill''.

We present our ablation results on the four model variations in Table~\ref{tab:ablation-results}.

\noindent \textbf{Contrastive vs.\ Ranking loss}. We first define the max margin loss following notations in the image-sentence level contrastive loss (Eq.~\ref{eq:obj_contrast}) as
\begin{equation}
\small
\label{eq:margin}
\mathcal{L}_{IS}(X_i, Y_j) = \mathbb{E}_{\mathcal{N}(X_i) \in X} \left[h(X_i, Y_j)\right]
\end{equation}
\begin{equation*}
\resizebox{.45\textwidth}{!} 
{
 $h(X_i, Y_j) = \sum_{i'\in \mathcal{N}(X_i)} \max\{0, m - S(X_i, Y_j) + S(X_i', Y_j)\} $,
}
\end{equation*}
where $m$ is the margin. In our experiment, we fix $m = 0.05$.

 We observe that NCE loss substantially outperforms the standard max margin loss by +6.2\%/+3.7\% on Flickr30K Entities and ReferItGame, respectively. These results suggest the effectiveness of contrastive learning, as also demonstrated in the concurrent work~\cite{gupta2020contrastive}.

\begin{table}[t]
  \centering
  \begin{adjustbox}{width=1\columnwidth}
  \begin{tabular}{lcc}
    \toprule
    Method     &  Flickr30K ACC (\%) & ReferItGame ACC (\%)      \\
    \midrule
    Max Margin  & 42.11 & 22.94 \\ 
    NCE         & 48.35 & 26.63 \\ 
    Distill     & 45.05 & 17.25 \\ 
    NCE+Distill & \textbf{50.96} & \textbf{27.59} \\
    \bottomrule
  \end{tabular}
  \end{adjustbox}\vspace{-0.5em}
  \caption{Ablation our proposed methods on Flickr30K and ReferItGame. All models use Res101 CC as backbone and models with distillation use IRV2 OI as Detector$_{K}$.}
  \label{tab:ablation-results}\vspace{-1.5em}
\end{table}

\noindent \textbf{Effects of Knowledge Distillation}.
Our full model combining both region-phrase and image-sentence matching  brings further improvements over both the NCE only model and the Distill only model. We conjecture that NCE and Distill provide complementary information for phrase grounding.
Specifically, Distill helps learn the extent of objects better, distinguishing parts frequently co-occurred in the same image, such as ``person'' and ``face''. NCE helps to learn finer-grained attributes, such as ``striped shirt'' and ``blue shirt'', as well as concepts not covered by detector classes.

To understand where the accuracy improvement of the distillation method comes from, we compute per phrase accuracy on frequent phrases that are mapped to one of the classes in the Open Images Dataset. The results can be found in the supplementary materials. We also perform qualitative analysis with  side-by-side heatmaps of region-phrase matching scores on such mapped phrases, shown in Figure~\ref{fig:compare}. Our full model (NCE+Distill) can better localize objects corresponding to the query phrase. 

\noindent \textbf{Generalization to Different Backbones}.
We vary the object detectors used by our model and present the results in Table~\ref{tab:distillation-results}. As the knowledge gap between visual backbone and external detector becomes smaller, e.g., the visual backbones are trained on tasks that involve finer-grained object and attribute annotations, the effects of distillation become less predominant. Nevertheless, our method can consistently improve performance for all detectors.

\begin{table}[t]
  \centering
  \begin{adjustbox}{width=1\columnwidth}
  \begin{tabular}{llcc}
    \toprule
     Backbone  & Detector$_{k}$ & NCE ACC (\%) & NCE+Distill ACC (\%)      \\
    \midrule
    Flickr30k & & & \\
    VGG16 PV  & VGG16 CC & 35.96 & 40.38 (+4.42) \\ 
    Res101 CC  & IRV2 OI & 48.35 & 50.96 (+2.61) \\ 
    Res101 VG  & IRV2 OI & 51.31 & 53.10 (+1.79) \\ 
    \midrule
    ReferItGame & & & \\
    VGG16 PV  & VGG16 CC & 23.56 & 24.52 (+0.96) \\ 
    Res101 CC & IRV2 OI  & 26.63 & 27.59 (+0.96) \\
    Res101 VG  & IRV2 OI & 36.65 & 38.39 (+1.74) \\ 
    \bottomrule
  \end{tabular}
  \end{adjustbox}\vspace{-0.5em}
  \caption{Results of our proposed distillation methods on Flickr30K and ReferItGame using various backbones and detectors providing external knowledge. “Backbone” denotes the visual backbone used to extract region features. Detector$_{K}$ denotes the detector that provides external knowledge. Dataset notations: PV=PASCAL VOC, CC=COCO, OI=Open Images, and VG=Visual Genome.}\label{tab:distillation-results}\vspace{-1.5em}
\end{table}

\section{Conclusion, Limitation and Future Work}
\label{conclusion}
In this paper, we presented a novel contrastive learning framework for weakly supervised visual phrase grounding. The key idea of our method is to learn a score function measuring the similarity between region-phrase pairs, distilled from object detection outputs and further supervised by image-sentence pairs. Once learned, this score function can be used for visual grounding without the need of object detectors at test time. While conceptually simple, our method demonstrated strong results on major benchmarks, surpassing state-of-the-art methods that use expensive object detectors. Our work offers a principled approach to leverage object information, as well as an efficient method for weakly supervised grounding. %We believe that our work provides a step towards bridge the gap between vision and language.

\noindent \textbf{Limitation and Future Work}. A main limitation of our work is the need of a generic object detector to cover most of the object classes during training. While object detection is currently more mature than visual grounding, learning open-set object detectors from large scale web data and further bridging the gap between object detection and visual grounding is a promising direction. Moving forward, we also plan to extend our method to solve other multi-modal grounding tasks, including video grounding~\cite{anne2017localizing}.

\bibliographystyle{ieee_fullname}
\bibliography{cvpr2020}

\end{document}